\documentclass{article}

\usepackage{arxiv}

\usepackage[utf8]{inputenc} 
\usepackage[T1]{fontenc}    
\usepackage{hyperref}       
\usepackage{url}            
\usepackage{booktabs}       
\usepackage{amsmath}
\usepackage{amsfonts}       
\usepackage{nicefrac}       
\usepackage{microtype}      
\usepackage{graphicx}
\usepackage{natbib}\setcitestyle{round}
\usepackage{doi}
\usepackage{caption}
\usepackage{subcaption}

\title{Robust volatility updates for Hierarchical Gaussian Filtering}

\date{} 					

\author{%
  \protect\href{https://orcid.org/0000-0003-4079-5453}{\protect\includegraphics[scale=0.06]{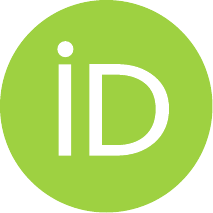}\hspace{1mm}Christoph Mathys}\textsuperscript{1}\thanks{\texttt{chmathys@cc.au.dk}}\quad
  Nicolas Legrand\textsuperscript{2}\quad
  Peter Thestrup Waade\textsuperscript{3}\quad
  Nace Mikus\textsuperscript{1}\quad
  Lilian Aline Weber\textsuperscript{4} \\[0.6em]
  \small\textsuperscript{1}Interacting Minds Centre, Aarhus University, 8000 Aarhus C, Denmark \\
  \small\textsuperscript{2}Centre for Humanities Computing, Aarhus University, 8000 Aarhus C, Denmark \\
  \small\textsuperscript{3}Translational Neuromodeling Unit, University of Zurich \& ETH Zurich, 8032 Zurich, Switzerland \\
  \small\textsuperscript{4}Institute for Cognitive Science, University of Osnabrück, 49090 Osnabrück, Germany
}

\renewcommand{\headeright}{Technical Report}
\renewcommand{\undertitle}{Technical Report}
\renewcommand{\shorttitle}{Robust volatility updates for HGF}

\hypersetup{
pdftitle={Robust volatility updates for Hierarchical Gaussian Filtering},
pdfsubject={cs.LG, cs.NE, q-bio.NC},
pdfauthor={Christoph Mathys, Nicolas Legrand, Peter Thestrup Waade, Nace Mikus, Lilian Aline Weber},
pdfkeywords={hierarchical Gaussian filtering, variational inference, quadratic approximation, Lambert W function, volatility estimation, computational psychiatry},
}

\begin{document}
\maketitle

\begin{abstract}
Hierarchical Gaussian Filtering (HGF) networks allow for efficient updating of posterior distributions (beliefs) about hidden states of an agent's environment. HGF parent nodes can target the mean or variance of their children. New information entering at input nodes leads to a cascade of belief updates across the network according to one-step update equations for each node's mean and precision (inverse variance). However, the original form of the update equations for variance-targeting parents (volatility coupling) can in some regions of parameter space lead to negative posterior precision, a logical impossibility which causes the updating algorithm to terminate with an error. In this report, we introduce a modified quadratic approximation to the variational energy of volatility-coupled nodes that avoids negative posterior precision. The key idea is to interpolate between two quadratic expansions of the variational energy: one at the prior prediction and one at a second mode whose location is obtained in closed form via the Lambert~$W$ function. The resulting update equations are robust across the entire parameter space and faithfully track the variational posterior even for large prediction errors.
\end{abstract}

\keywords{hierarchical Gaussian filtering \and variational inference \and quadratic approximation \and Lambert~$W$ function \and volatility estimation \and computational psychiatry}

\section{Introduction}
\label{sec:intro}

Hierarchical Gaussian filters \citep[HGFs;][]{mathys_bayesian_2011, mathys_uncertainty_2014} are networks of nodes that allow for modelling an agent's inference on states of its environment to which it has only indirect access through observations or measurements made by input nodes. They rest on a variational approximation to Bayesian inference in a hierarchical generative model where higher levels control the volatility (rate of change) of lower levels, yielding one-step update equations in the form of precision-weighted prediction errors. HGFs have found wide application in computational psychiatry and cognitive neuroscience, where they serve as individual-level models of learning, perception, and belief updating whose parameters can be estimated from behavioural data \citep[e.g.,][]{iglesias_hierarchical_2013, diaconescu_inferring_2014, powers_pavlovian_2017, lawson_adults_2017, mikus_computational_2024}. They have also been applied to financial time series \citep{mathys_bayesian_2011} and to active inference \citep{mathys_hierarchical_2020, waade_inferring_2021}. A freely available software implementation is provided in the Julia package \texttt{HierarchicalGaussianFiltering.jl}\footnote{\url{https://github.com/ComputationalPsychiatry/HierarchicalGaussianFiltering.jl}}, the Python package \texttt{pyhgf}\footnote{\url{https://github.com/ComputationalPsychiatry/pyhgf}}, and the MATLAB HGF Toolbox\footnote{\url{https://github.com/ComputationalPsychiatry/hgf-toolbox}}, all distributed as part of TAPAS \citep{frassle_tapas_2021}. HGFs were recently generalized to include nonlinear value coupling between parent and child nodes alongside the existing volatility coupling \citep{weber_generalized_2025}, enabling the construction of predictive coding hierarchies and deep artificial neural networks within the HGF framework.

Accordingly, there are two kinds of update equations that determine how information flows through an HGF network: value updates and volatility updates. Both consist of a pair of update equations, one for the mean, the other for the precision of the Gaussian that describes the filter's posterior distribution of a node, and in both cases the mean updates are precision-weighted prediction errors. While three of these four updates are unproblematic (mean and precision updates for value coupling, and the mean update for volatility coupling), the precision update for volatility coupling can sometimes give rise to negative posterior precision, which is a logical impossibility because the variance of a Gaussian has to be positive.

Negative posterior precision can arise because for some combinations of node states and parameter values, the quadratic approximation to the node's variational energy is inappropriate. In this report, we first briefly revisit the way the HGF update equations are derived. We then give a detailed explanation of how and why negative posterior precision can arise with the original derivation. Armed with this understanding, we are then in a position to find a modified approximation which not only solves the problem of negative posterior precision but also preserves the main features of the original approach: quadratic approximation to the variational energy, volatility prediction errors, and simple updates on the basis of precision-weighted prediction errors. The new update equations make HGF networks robust across the entire expanse of parameter space, allowing for a wider range of models, and they remove the need for costly boundary checks during parameter estimation.

\section{Theory}
\label{sec:theory}

\subsection{Inference on HGF node values under the mean field approximation}

Under the mean field approximation, the posterior probability distribution $q$ on the value of a node $x$ in a network is the normalized exponential of the node's variational energy $I$ \citep[for details see][Appendix B]{mathys_hierarchical_2012}:

\begin{equation}
    \label{eq:varpost}
    q(x) = \frac{1}{\mathcal{Z}} \exp(I(x)) \quad \text{with } \mathcal{Z} := \int_{-\infty}^{\infty} \exp(I(x)) \mathrm{d}x    
\end{equation}

This means that the approximate posterior $q$ is Gaussian if and only if the variational energy $I$ is quadratic.

\subsection{Variational energy function of volatility coupling}

An HGF state node $x$ which is volatility parent of $x_\dagger$ and in turn has volatility parent $x_\ddagger$ has the following variational energy \citep{mathys_bayesian_2011, mathys_uncertainty_2014}, where $t$ denotes the time elapsed since the previous update: 

\begin{equation}
    I(x) = - \frac{1}{2} \log \left( \sigma_\dagger^0 + t \exp(\kappa_\dagger x + \omega_\dagger) \right) - \frac{1}{2} \frac{\sigma_\dagger + \left( \mu_\dagger - \hat{\mu}_\dagger \right)^2}{\sigma_\dagger^0 + t \exp(\kappa_\dagger x + \omega_\dagger)} - \frac{1}{2} \frac{1}{\sigma^0 + t \exp(\kappa \mu_\ddagger^0 + \omega)} \left( x - \hat{\mu} \right)^2,
\end{equation}

The parameters of the variational energy $I$ are described in Table \ref{tab:ve_params}. This notation is based on \citet{mathys_uncertainty_2014}, with slight adjustments for better readability.

\begin{table}
    \centering
    \caption{Parameters of the variational energy}
    \begin{tabular}{ll}
    \toprule
    Name & Description \\[0.2em]
    \midrule
    $t$ & Time elapsed since the previous update\\[0.2em]
    $\sigma_\dagger^0$ & Previous $\sigma$ of child\\[0.2em]
    $\kappa_\dagger$ & $\kappa$ of child\\[0.2em]
    $\omega_\dagger$ & $\omega$ of child\\[0.2em]
    $\sigma_\dagger$ & $\sigma$ of child\\[0.2em]
    $\mu_\dagger$ & $\mu$ of child\\[0.2em]
    $\hat{\mu}_\dagger$ & Predicted $\mu$ of child\\[0.2em]
    $\sigma^0$ & Previous $\sigma$\\[0.2em]
    $\kappa$ & $\kappa$ of current node\\[0.2em]
    $\omega$ & $\omega$ of current node\\[0.2em]
    $\mu_\ddagger^0$ & Previous $\mu$ of parent\\[0.2em]
    $\hat{\mu}$ & Predicted $\mu$\\[0.2em]
    \bottomrule
    \end{tabular}
    \label{tab:ve_params}
\end{table}

\subsection{Canonical parameterization of the variational energy}

For our current purposes, it is useful to introduce a parameterization of the variational energy that uses only three parameters but retains all properties that we are concerned with here. We call this the \emph{canonical variational energy function} $J$.

We obtain it by setting $t$ to one and letting all $\mu$'s and $\omega$'s be zero and letting all $\sigma$'s and $\kappa$'s be one, with three exceptions. First, $\sigma_\dagger^0$ remains as a parameter and will be written as $\alpha$ for simplicity. Second, the posterior total uncertainty about the child $\sigma_\dagger + \left( \mu_\dagger - \hat{\mu}_\dagger \right)^2$ is taken as a single parameter $\beta$. Finally, $\hat{\mu}$ remains and is called $\gamma$.

\begin{equation}
    \label{eq:J}
    J(x) := - \frac{1}{2} \log \left( \alpha + e^x \right) - \frac{1}{2} \frac{\beta}{\alpha + e^x} - \frac{1}{4} (x - \gamma)^2 \quad \text{with } \alpha, \beta > 0, \gamma \in \mathbb{R}
\end{equation}

\subsection{Quadratic approximation in the original HGF formulation}

Since $J$ is not quadratic, the approximate posterior implied by it via Eq.~\ref{eq:varpost} is not Gaussian. To obtain a Gaussian approximate posterior, we need to find a quadratic approximation to $J$. In the original HGF formulation \citep{mathys_bayesian_2011}, this is achieved by expanding $J$ to second order at $\gamma$.

\begin{equation}
    \hat{J}(x) := J(\gamma) + J'(\gamma) (x - \gamma) + \frac{1}{2} J''(\gamma) (x - \gamma)^2.
\end{equation}

This yields a Gaussian posterior with mean $\mu_J$ and precision (inverse variance) $\pi_J$:

\begin{equation}
    \label{eq:q}
    q(x) = \sqrt{\frac{\pi_J}{2 \pi}} \exp \left(- \frac{\pi_J}{2} \left(x - \mu_J \right)^2 \right)
\end{equation}

In the calculation of $\mu_J$ and $\pi_J$, we need the first and second derivatives, $J'$ and $J''$, of $J$:

\begin{align}
    J'(x) =& \frac{1}{2} \frac{e^x}{\alpha + e^x} \left( \frac{\beta}{\alpha + e^x} - 1 \right) - \frac{1}{2} (x - \gamma) \\[2em]
    J''(x) =& - \frac{1}{2} \frac{e^x}{\alpha + e^x} \left( \frac{\alpha}{\alpha + e^x} + \frac{(e^x - \alpha) \beta}{(\alpha + e^x)^2} \right) - \frac{1}{2} \nonumber\\ =& - \frac{1}{2} \frac{e^x}{\alpha + e^x} \left( \frac{e^x}{\alpha + e^x} + \frac{e^x - \alpha}{\alpha + e^x} \left( \frac{\beta}{\alpha + e^x} - 1 \right) \right) - \frac{1}{2}.
\end{align}

We simplify the notation by defining

\begin{align}
    w(x) :=& \frac{e^x}{\alpha + e^x} \\[2em]
    \delta(x) :=& \frac{\beta}{\alpha + e^x} - 1,
\end{align}

which gives us

\begin{align}
    J'(x) =& \frac{1}{2} w(x) \delta(x) - \frac{1}{2} (x - \gamma) \\[2em]
    J''(x) =& - \frac{1}{2} w(x) \left( w(x) + (2w(x) - 1) \delta(x) \right) - \frac{1}{2}.
\end{align}

In these terms, the original HGF update equations \citep[cf. ][]{mathys_bayesian_2011, mathys_uncertainty_2014} read

\begin{align}
    \label{eq:piJ}
    \pi_J =& - J''(\gamma) = \frac{1}{2} w(\gamma) \left( w(\gamma) + (2w(\gamma) - 1) \delta(\gamma) \right) + \frac{1}{2} \\[2em]
    \mu_J =& \gamma - \frac{J'(\gamma)}{J''(\gamma)} = \gamma + \frac{w(\gamma)}{2 \pi_J} \delta(\gamma),
\end{align}

where $\pi_J$ and $\mu_J$ are the sufficient statistics of the Gaussian approximate posterior $q$ of Eq.~\ref{eq:q}.

\subsection{Negative posterior precision}

The main problem with this original approach is that the posterior precision $\pi_J$ can under certain circumstances become negative according to Eq.~\ref{eq:piJ}. Negative precision is, however, logically impossible and can occur only algorithmically at points where the approximation breaks down.

For an interactive graphical illustration of the phenomenon of negative posterior precision, a Pluto notebook \citep{plas_fonspplutojl_2024} is available at \url{https://github.com/ilabcode/UnboundedHGF.jl}. Starting with values of $\alpha = \beta = 1$ and varying $\gamma$, we see that when the previous variance $\alpha$ of (that is, uncertainty about) the child node is on a similar scale to the total posterior uncertainty $\beta$ about the child, then negative posterior precision does not arise and the original quadratic approximation is unproblematic. However, if we now reduce $\alpha$ to $0.005$, reflecting a situation where a very large prediction error leads to a total posterior uncertainty much greater than the previous uncertainty, we now see values of $\gamma$ for which the posterior precision is negative, indicated by the flipping around of the quadratic approximation's parabola so that it opens above instead of below.

\begin{figure}
    \centering
    \begin{subfigure}[t]{0.4\linewidth}
        \centering
        \includegraphics[width=\linewidth]{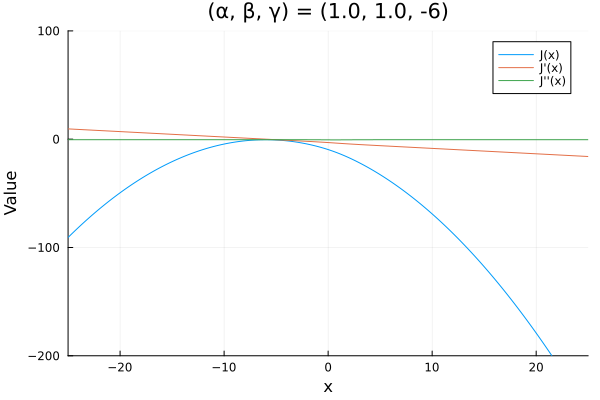}
        \caption{Scenario 1: Variational energy and derivatives}
        \label{fig:derivs1}        
    \end{subfigure}
    \hspace{4em}
    \begin{subfigure}[t]{0.4\linewidth}
        \centering
        \includegraphics[width=\linewidth]{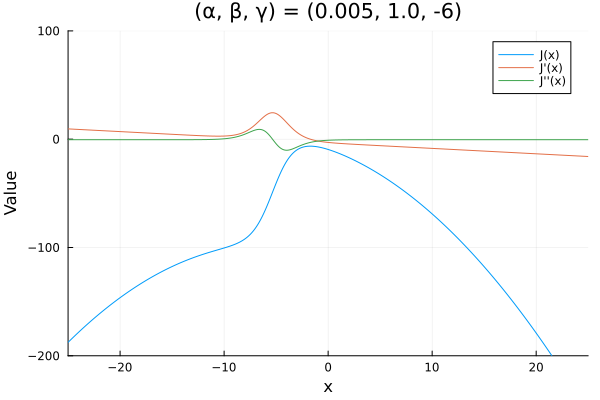}
        \caption{Scenario 2: Variational energy and derivatives}
        \label{fig:derivs2}        
    \end{subfigure}
    \par\bigskip
    \begin{subfigure}[b]{0.4\linewidth}
        \centering
        \includegraphics[width=\linewidth]{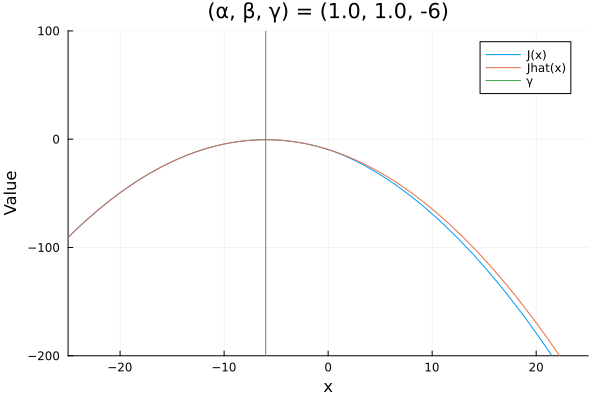}
        \caption{Scenario 1: Quadratic approximation $\hat{J}$}
        \label{fig:approx1}        
    \end{subfigure}
    \hspace{4em}
    \begin{subfigure}[b]{0.4\linewidth}
        \centering
        \includegraphics[width=\linewidth]{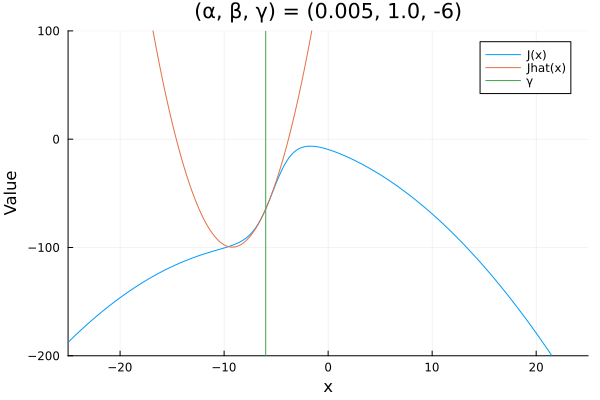}
        \caption{Scenario 2: Quadratic approximation $\hat{J}$}
        \label{fig:approx2}        
    \end{subfigure}
    \caption{Two scenarios for the original quadratic approximation. In Scenario 1, $\alpha$ and $\beta$ are of comparable magnitude, leading to an approximately quadratic variational energy $J$. However, in Scenario 2, $\alpha$ has been reduced $200$-fold, leading to a convex region in the variational energy. If the expansion point $\gamma$ now happens to fall into this region, the quadratic approximation fails.}
    \label{fig:origapprox}
\end{figure}

Fig.~\ref{fig:origapprox} shows two scenarios for $(\beta, \gamma) = (1, 6)$. Scenario 1 has $\alpha = 1$ (left-hand panels \ref{fig:derivs1} and \ref{fig:approx1}) while Scenario 2 has $\alpha = 0.005$ (right-hand panels \ref{fig:derivs2} and \ref{fig:approx2}). In Scenario 1, the first derivative is approximately a straight line with constant slope, and the second derivative is accordingly approximately constant (Fig.~\ref{fig:derivs1}). The quadratic approximation $\hat{J}$ is therefore almost exact (Fig.~\ref{fig:approx1}). In Scenario 2, however, the variational energy has a convex region where the second derivative goes above $0$ (Fig.~\ref{fig:derivs2}) and the quadratic approximation fails (Fig.~\ref{fig:approx2}).

If negative posterior precision occurs on at least one occasion in the course of filtering a time series, then the chosen parameter set is out of bounds, and another has to be used where posterior precision never turns negative. This places boundaries in parameter space beyond which the algorithm cannot go. While in many practical cases, this is only a minor nuisance, it would still be desirable to have a quadratic approximation that works for all parameter choices in that it never produces negative posterior precision.

To find such an improved approximation, we take a closer look at $J$ and its components and their derivatives. Since posterior precision is the negative of the second derivative, it is especially important to understand how and where this can turn positive. We go about finding an improved approximation in two steps. First, we eliminate the problem of negative posterior precision. Once this is done, we ensure that the quadratic approximation always leads to a Gaussian which closely approximates the non-Gaussian variational posterior.

\section{Results}

\subsection{Ensuring positive posterior precision}

\begin{figure}
    \centering
    \includegraphics[width=0.85\linewidth]{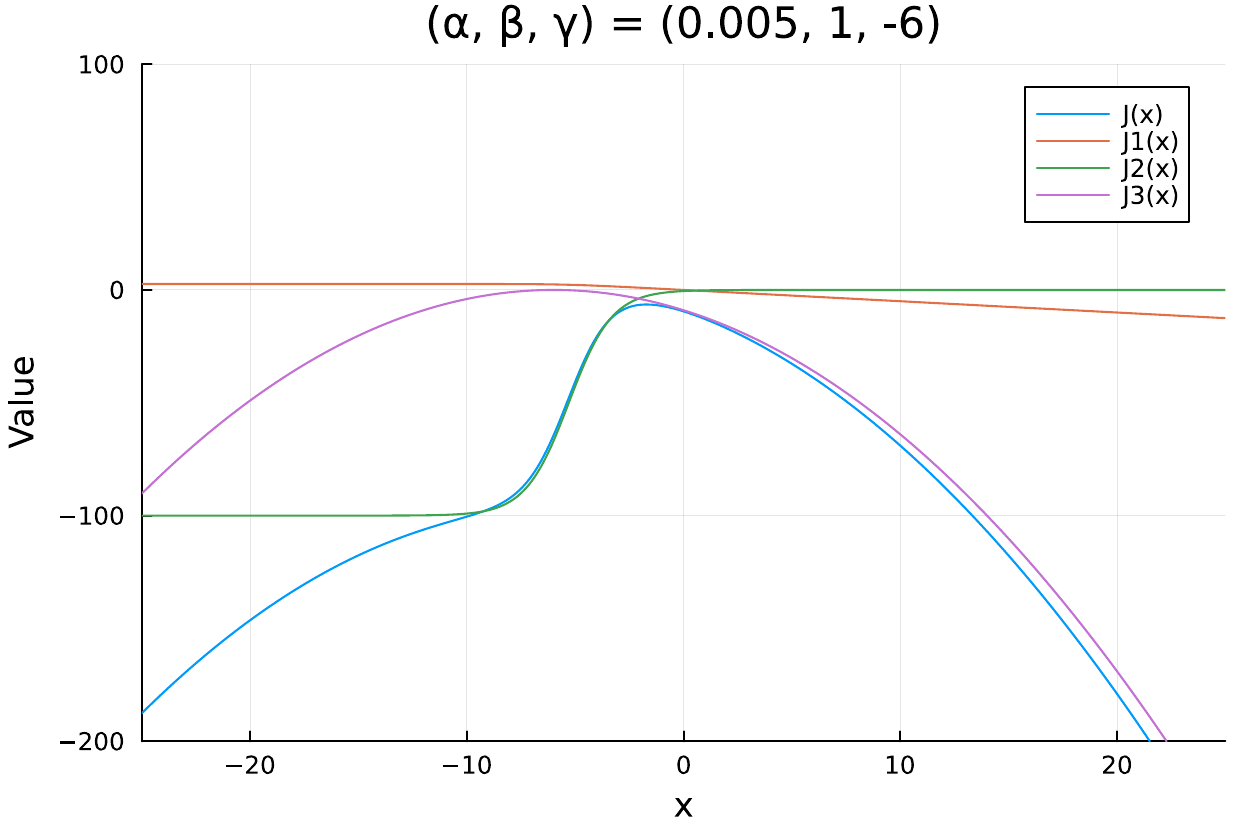}
    \caption{The three components of the canonical variational energy $J(x) = J_1(x) + J_2(x) + J_3(x)$ for $(\alpha, \beta, \gamma) = (0.005, 1, -6)$. $J_1$ is concave everywhere. $J_2$ makes a sigmoid-like transition from $-\beta/2\alpha$ to $0$ in a narrow interval around $x = \log\alpha$, with a convex region in this transition zone. $J_3$ is a downward parabola centred at $\gamma$.}
    \label{fig:Jcomp}
\end{figure}

In order to understand how negative posterior precision can arise in our original approach, we look at the three summands of the variational energy $J$ from Eq.~\ref{eq:J} separately (Fig.~\ref{fig:Jcomp}):

\begin{align}
    J_1(x) :=& - \frac{1}{2} \log \left( \alpha + e^x \right)\\[1.2em]
    J_2(x) :=& -\frac{1}{2} \frac{\beta}{\alpha + e^x}\\[1.2em]
    J_3(x) :=& - \frac{1}{4} (x - \gamma)^2
\end{align}

$J_1$ is concave, approaching the constant $-\frac{1}{2} \log(\alpha)$ for small $x$ and the linear $-\frac{1}{2}x$ for large $x$. Its second derivative is negative everywhere, so it cannot contribute to negative posterior precision. $J_2$ on the other hand is partly convex and can therefore give rise to regions of negative posterior precision. However, it is clear that this can only happen in a narrowly confined interval of $x$. To the left of this, $J_2$ effectively amounts to the constant $-\beta/2\alpha$, and to the right of this, to $0.$  In Fig.~\ref{fig:Jcomp}, the transition region where $J_2$ changes from $-\beta/2\alpha$ to $0$ is clearly visible. Finally, $J_3$ is a quadratic function, which implies that its second derivative is constant at $-1/2$, which means that $J_3$, like $J_1$, cannot give rise to negative posterior precision.

The full picture thus is that $J$ is quadratic both to the left and to the right of $J_2$'s transition region, and that the only difference between these two regions is the linear term $-\frac{1}{2}x$ contributed by $J_1$ which is present on the right but not the left. Crucially, this term does not affect the second derivative, meaning that posterior precision is the same whether calculated by expanding $J$ to second order on the right or on the left. The only problematic region is in the middle, where $J_2$ makes the transition from $-\beta/2\alpha$ to $0.$

With the problem laid out in this manner, there is now an obvious solution to negative posterior precision as a result of the quadratic approximation: we disregard the offending term $J_2$ when calculating precision. This leaves us with the \emph{concave canonical variational energy} $K$:

\begin{align}
    K(x) :=& J_1(x) + J_3(x) \nonumber\\
    =& - \frac{1}{2} \log \left( \alpha + e^x \right) - \frac{1}{4} (x - \gamma)^2
\end{align}

The first and second derivatives of $K$ are:

\begin{align}
    K'(x) =& - \frac{1}{2} \frac{e^x}{\alpha + e^x} - \frac{1}{2} (x - \gamma)\nonumber\\[1em] =& - \frac{1}{2} w(x) - \frac{1}{2} (x - \gamma)\\[2em]
    K''(x) =& - \frac{1}{2} \frac{\alpha e^x}{\left( \alpha + e^x \right)^2} - \frac{1}{2}\nonumber\\[1em] =& - \frac{1}{2} w(x) (1 - w(x)) - \frac{1}{2}
\end{align}

Since $\alpha$ is positive, the second derivative is negative everywhere, and negative posterior precision cannot occur. This enables us to construct a quadratic approximation $L_1$ where the precision $\pi_{L_1}$ is $-K''(\gamma)$ and the mean $\mu_{L_1}$ remains as before except that it uses the precision according to its new definition $\pi_{L_1}$

\begin{equation}
    L_1(x) = - \frac{\pi_{L_1}}{2} \left(x - \mu_{L_1} \right)^2
\end{equation}

with

\begin{align}
    \pi_{L_1} =& - K''(\gamma) = \frac{1}{2} w(\gamma)(1 - w(\gamma)) + \frac{1}{2} \\[2em]
    \mu_{L_1} =& \gamma - \frac{J'(\gamma)}{K''(\gamma)} = \gamma + \frac{w(\gamma)}{2 \pi_{L_1}} \delta(\gamma)
\end{align}

\subsection{The case of two local maxima in the variational energy}

\begin{figure}
    \centering
    \includegraphics[width=0.85\linewidth]{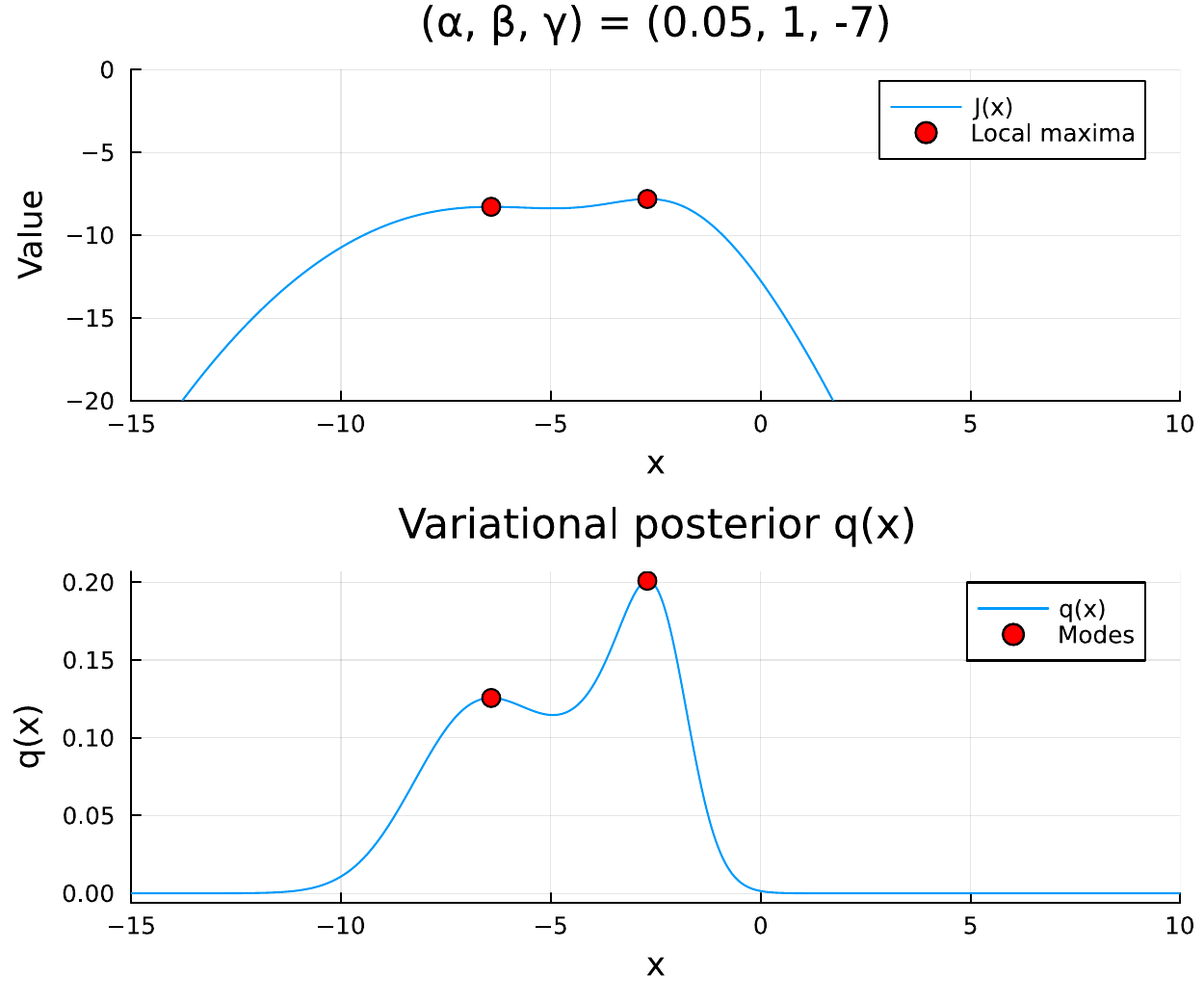}
    \caption{Bimodal variational energy and posterior for $(\alpha, \beta, \gamma) = (0.05, 1, -7)$. Top: the variational energy $J(x)$ has two local maxima (red dots), separated by a valley caused by the convex region of~$J_2$. Bottom: the corresponding variational posterior $q(x) \propto \exp(J(x))$ is bimodal. A single Gaussian cannot capture both modes; the two-expansion approach of Section~3.2 addresses this.}
    \label{fig:bimodal}
\end{figure}

For some combinations of $\alpha$, $\beta$, and $\gamma$, the term $J_2$ does not only give rise to a convex region in the variational energy $J$ but also to a second local maximum (Fig.~\ref{fig:bimodal}). In these cases, the variational posterior, obtained by exponentiation and normalization of the variational energy, becomes bimodal. This forces us to take special care in order to ensure that our quadratic approximation to the variational energy produces a Gaussian posterior which reflects the variational posterior as faithfully as possible. To handle this, we construct a second quadratic expansion $L_2$ near the second mode and combine $L_1$ and $L_2$ into a single Gaussian via softmax-weighted mixture moment matching.

\paragraph{Finding the second mode via the Lambert $W$ function.}

The stationary points of $J$ satisfy $J'(x) = 0$, i.e.,
\begin{equation}
    \frac{1}{2} w(x) \, \delta(x) = \frac{1}{2} (x - \gamma).
\end{equation}
This transcendental equation cannot be solved in closed form. However, in the limit $\alpha \to 0$ (i.e., the regime where the second mode appears and where the original approximation fails), we have $w(x) \to 1$ and $\delta(x) \to \beta \, e^{-x} - 1$, so the mode equation simplifies to
\begin{equation}
    \beta \, e^{-x} = 1 + x - \gamma.
\end{equation}
Substituting $u := 1 + x - \gamma$ gives $u \, e^{u} = \beta \, e^{1 - \gamma}$, which is solved by the principal branch $W_0$ of the Lambert~$W$ function \citep{corless_lambert_1996}:
\begin{equation}
    \label{eq:lambertw_mode}
    x^* = \gamma - 1 + W_0\!\left(\beta \, e^{1 - \gamma}\right).
\end{equation}
Since $\beta \, e^{1-\gamma} > 0$, the solution is unique and real. The Lambert $W_0$ function can be evaluated to machine precision in a few iterations of Halley's method. This gives us an analytic approximation to the second mode that is accurate whenever $\alpha$ is small relative to $e^{x^*}$, which is precisely the regime where the bimodal case arises and where the original quadratic approximation fails.

\paragraph{Second quadratic expansion.}

We construct the second quadratic expansion $L_2$ at $x^*$. Because $x^*$ approximates a local maximum of $J$, the second derivative $J''(x^*)$ is negative there, so the expansion yields positive precision:
\begin{equation}
    L_2(x) = - \frac{\pi_{L_2}}{2} \left(x - \mu_{L_2} \right)^2    
\end{equation}
with
\begin{align}
    \pi_{L_2} =& - J''(x^*) \nonumber\\[1em]
    =& \frac{1}{2} w(x^*) \left( w(x^*) + (2w(x^*) - 1) \, \delta(x^*) \right) + \frac{1}{2} \\[2em]
    \mu_{L_2} =& x^* - \frac{J'(x^*)}{J''(x^*)} \nonumber\\[1em]
    =& x^* + \frac{1}{\pi_{L_2}} \left( \frac{1}{2} w(x^*) \, \delta(x^*) - \frac{1}{2}(x^* - \gamma) \right).
\end{align}
We use the full second derivative $J''$ (not $K''$) for the expansion at the approximate mode, because the curvature is concave there. The residual $J'(x^*)$, generally small because $x^*$ is close to the true mode, is corrected by the Newton step $-J'(x^*)/J''(x^*)$. In the unlikely case that $\pi_{L_2} \leq 0$ (which can occur when $x^*$ falls far from the actual mode), we fall back to $\pi_{L_2} = -K''(x^*)$, which is always positive.

\paragraph{Softmax blending.}

We weight the two expansions according to the variational energy $J$ evaluated at their respective means:
\begin{equation}
    \label{eq:softmax}
    b = \frac{1}{1 + \exp\!\left(J(\mu_{L_1}) - J(\mu_{L_2})\right)}.
\end{equation}
This assigns higher weight to whichever expansion sits at a point of higher variational energy, i.e., closer to the dominant mode. When the variational posterior is unimodal near $\gamma$, we have $J(\mu_{L_1}) > J(\mu_{L_2})$ and $b \approx 0$, recovering $L_1$. When the second mode dominates, $J(\mu_{L_2}) > J(\mu_{L_1})$ and $b \approx 1$.

\paragraph{Gaussian mixture moment matching.}

We treat the two expansions as components of a two-component Gaussian mixture and compute the moment-matched mean and variance:
\begin{align}
    \label{eq:mixture_mu}
    \mu_L =& (1 - b) \, \mu_{L_1} + b \, \mu_{L_2} \\[1.5em]
    \label{eq:mixture_var}
    \sigma_L^2 =& \frac{1 - b}{\pi_{L_1}} + \frac{b}{\pi_{L_2}} + b(1-b)\left(\mu_{L_1} - \mu_{L_2}\right)^2 \\[1.5em]
    \pi_L =& \frac{1}{\sigma_L^2}.
\end{align}
The third term $b(1-b)(\mu_{L_1} - \mu_{L_2})^2$ is the between-component variance, which ensures that when the two expansion means are far apart and both carry substantial weight, the resulting Gaussian is appropriately wide. When $b \approx 0$ or $b \approx 1$, this term vanishes and we recover one of the individual expansions. This moment matching corresponds exactly to the mean and variance of the two-component mixture $q(x) = (1-b) \, \mathcal{N}(x;\, \mu_{L_1},\, \pi_{L_1}^{-1}) + b \, \mathcal{N}(x;\, \mu_{L_2},\, \pi_{L_2}^{-1})$, and the posterior precision $\pi_L$ is strictly positive by construction.

\subsection{Variational posterior and Gaussian approximation}

The (unnormalized) marginal posterior distribution under the mean field approximation is the exponential of the variational energy, which we can now compare visually with our Gaussian approximation.

In the Pluto notebook available at \url{https://github.com/ilabcode/UnboundedHGF.jl}, there are interactive figures that show the variational energy, $J$, our new approximation $L$, and its components $L_1$ and $L_2$ in logarithmic space, where the approximations are quadratic, as well as in native space as probability densities, where the approximations are Gaussian.

It is instructive to start a visual exploration of the approximation's behaviour with $\alpha$ and $\beta$ both set to 1 while varying $\gamma$ from -40 to +40. Throughout this range, $L_1$ offers an excellent approximation which in native probability space is almost indistinguishable from the variational posterior. As in the left-hand panels of Fig.~\ref{fig:origapprox}, this reflects a situation where the previous posterior uncertainty $\sigma_\dagger^0$ about the child node (i.e., $\alpha$) is on a similar scale as the total posterior uncertainty $\sigma_\dagger + \left( \mu_\dagger - \mu_\dagger^0 \right)^2$ about that node (i.e., $\beta$).

However, if we again reduce $\alpha$ to 0.005 while leaving $\beta$ at 1 and again vary $\gamma$, we see that in this scenario, where uncertainty was very low relative to the update resulting from a very large prediction error, the interpolation between $L_1$ and $L_2$ becomes relevant and ensures that our approximation $L$ keeps reflecting the variational posterior across the whole range of $\gamma$.

\subsection{Full equations}

We can now turn back to the fully parameterized form $I$ of the variational energy to give the full equations for HGF volatility prediction error updates \citep[cf.][Eqs 29--34]{mathys_bayesian_2011}:

\begin{align}
    \pi_{L_1} =& \hat{\pi} + \frac{\kappa_\dagger^2}{2} w_\dagger (1 - w_\dagger) \\[2em]
    \mu_{L_1} =& \hat{\mu} + \frac{\kappa_\dagger w_\dagger}{2 \pi_{L_1}} \delta_\dagger
\end{align}

with

\begin{align}
    \hat{\pi} :=& \frac{1}{\sigma^0 + t\exp(\kappa \mu_\ddagger^0 + \omega)} \\[1em]
    w_\dagger :=& \frac{t\exp(\kappa_\dagger \hat{\mu} + \omega_\dagger)}{\sigma_\dagger^0 + t\exp(\kappa_\dagger \hat{\mu} + \omega_\dagger)} \\[1em]
    \delta_\dagger :=& \frac{\sigma_\dagger + \left( \mu_\dagger - \hat{\mu}_\dagger \right)^2}{\sigma_\dagger^0 + t\exp \left( \kappa_\dagger \hat{\mu} + \omega_\dagger \right)} - 1
\end{align}

and

\begin{align}
    \pi_{L_2} =& \hat{\pi} + \frac{\kappa_\dagger^2}{2} w^* \left(w^* + (2 w^* - 1) \, \delta^*\right) \\[2em]
    \mu_{L_2} =& x^* + \frac{1}{\pi_{L_2}} \left( \frac{\kappa_\dagger}{2} w^* \, \delta^* - \hat{\pi}(x^* - \hat{\mu}) \right)
\end{align}

The expansion point $x^*$ is the approximate mode of the variational energy $I$ in the limit $\sigma_\dagger^0 \to 0$. It is found via the Lambert~$W$ function by first converting to the canonical variable $y = \log t + \kappa_\dagger x + \omega_\dagger$, in which the prior term of $I$ takes the form $-\frac{1}{2} \hat{\pi}_y (y - \gamma_c)^2$ with

\begin{align}
    \gamma_c :=& \log t + \kappa_\dagger \hat{\mu} + \omega_\dagger \\[1em]
    \hat{\pi}_y :=& \frac{\hat{\pi}}{\kappa_\dagger^2}.
\end{align}

The mode equation in $y$ then yields

\begin{equation}
    y^* = \gamma_c - \frac{1}{2\hat{\pi}_y} + W_0\!\left(\frac{\beta}{2\hat{\pi}_y} \, e^{1/(2\hat{\pi}_y) - \gamma_c}\right),
\end{equation}

where $\beta = \sigma_\dagger + (\mu_\dagger - \hat{\mu}_\dagger)^2$. Converting back to the native variable:

\begin{equation}
    x^* = \frac{y^* - \log t - \omega_\dagger}{\kappa_\dagger}.
\end{equation}

The quantities $w^*$ and $\delta^*$ are $w$ and $\delta$ evaluated at $x^*$ instead of $\hat{\mu}$:

\begin{align}
    w^* :=& \frac{t\exp(\kappa_\dagger x^* + \omega_\dagger)}{\sigma_\dagger^0 + t\exp(\kappa_\dagger x^* + \omega_\dagger)} \\[1em]
    \delta^* :=& \frac{\sigma_\dagger + (\mu_\dagger - \hat{\mu}_\dagger)^2}{\sigma_\dagger^0 + t\exp(\kappa_\dagger x^* + \omega_\dagger)} - 1.
\end{align}

The softmax weight and mixture moment matching follow Eqs.~\ref{eq:softmax}--\ref{eq:mixture_var}, with the canonical $J$ replaced by the full variational energy $I$:

\begin{equation}
    b = \frac{1}{1 + \exp\!\left(I(\mu_{L_1}) - I(\mu_{L_2})\right)}.
\end{equation}

\subsection{Comparative simulations}

We conducted four comparative simulations to verify that (i) the new update equations produce Gaussian posteriors that closely approximate the variational posterior, (ii) both methods produce valid filtering results under standard conditions, (iii) the new equations survive situations that cause the original equations to crash, and (iv) they thereby extend the usable region of parameter space. All simulations use a two-level continuous HGF. Code is available in the accompanying Julia package.\footnote{\url{https://github.com/ilabcode/UnboundedHGF.jl}}

\subsubsection{Approximation quality}

\begin{figure}[ht]
    \centering
    \includegraphics[width=\linewidth]{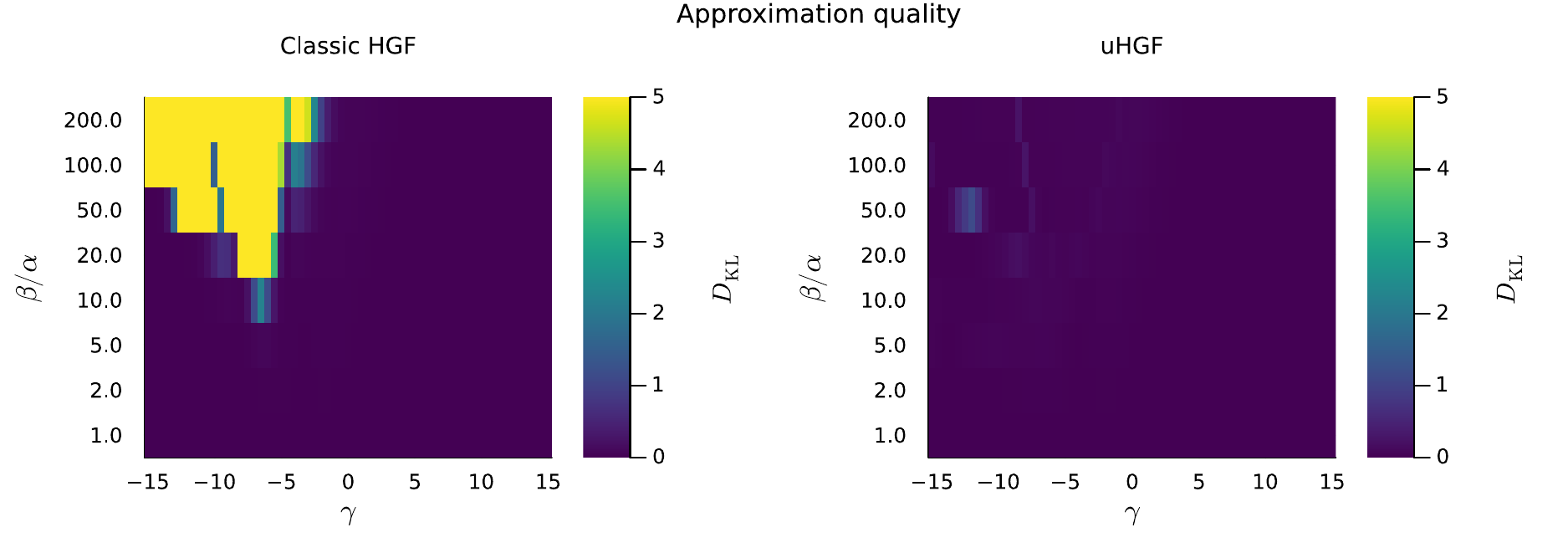}
    \caption{Simulation~1: KL divergence between the normalized variational posterior and the Gaussian approximation across the $\beta/\alpha$ vs.\ $\gamma$ parameter grid. Left: classic HGF (hot colours at negative $\gamma$ indicate failure). Right: uHGF (uniformly low divergence).}
    \label{fig:sim1}
\end{figure}

In the first simulation, we assessed how well the Gaussian posteriors implied by the classic and the new equations approximate the variational posterior, measured by the Kullback--Leibler (KL) divergence $D_\mathrm{KL}(p \| q)$, where $p$ is the normalized variational posterior (computed by numerical integration) and $q$ is the approximate Gaussian. We evaluated 488 parameter combinations spanning eight $\beta/\alpha$ ratios from 1 to 200 and 61 values of $\gamma$ from $-15$ to $15$, with $\alpha = 0.005$. Of these 488 points, the classic equations produced negative posterior precision (i.e., crashed) at 31 (6.4\%), whereas the new equations succeeded at all 488 (100\%). Where both methods succeeded, mean $D_\mathrm{KL}$ was 1.34 for the classic approximation and 0.023 for the new one --- an improvement of almost two orders of magnitude. This reflects the combined benefit of using the concave $K''$ for the first expansion's precision, the Lambert~$W$-guided second expansion point, and the softmax-weighted mixture moment matching. Fig.~\ref{fig:sim1} displays the KL divergences as heat maps: the classic map (left panel) contains a prominent band of failures at negative $\gamma$, whereas the uHGF map (right panel)---where uHGF stands for \emph{unbounded HGF}---is smooth and uniformly low. However, note that the classic HGF already delivered a very faithful approximation except for the combination of very large prediction errors ($\beta/\alpha > 10$) and an unfavourable, but limited, range of $\gamma$ values.

\subsubsection{Filtering under standard conditions}

\begin{figure}[ht]
    \centering
    \includegraphics[width=\linewidth]{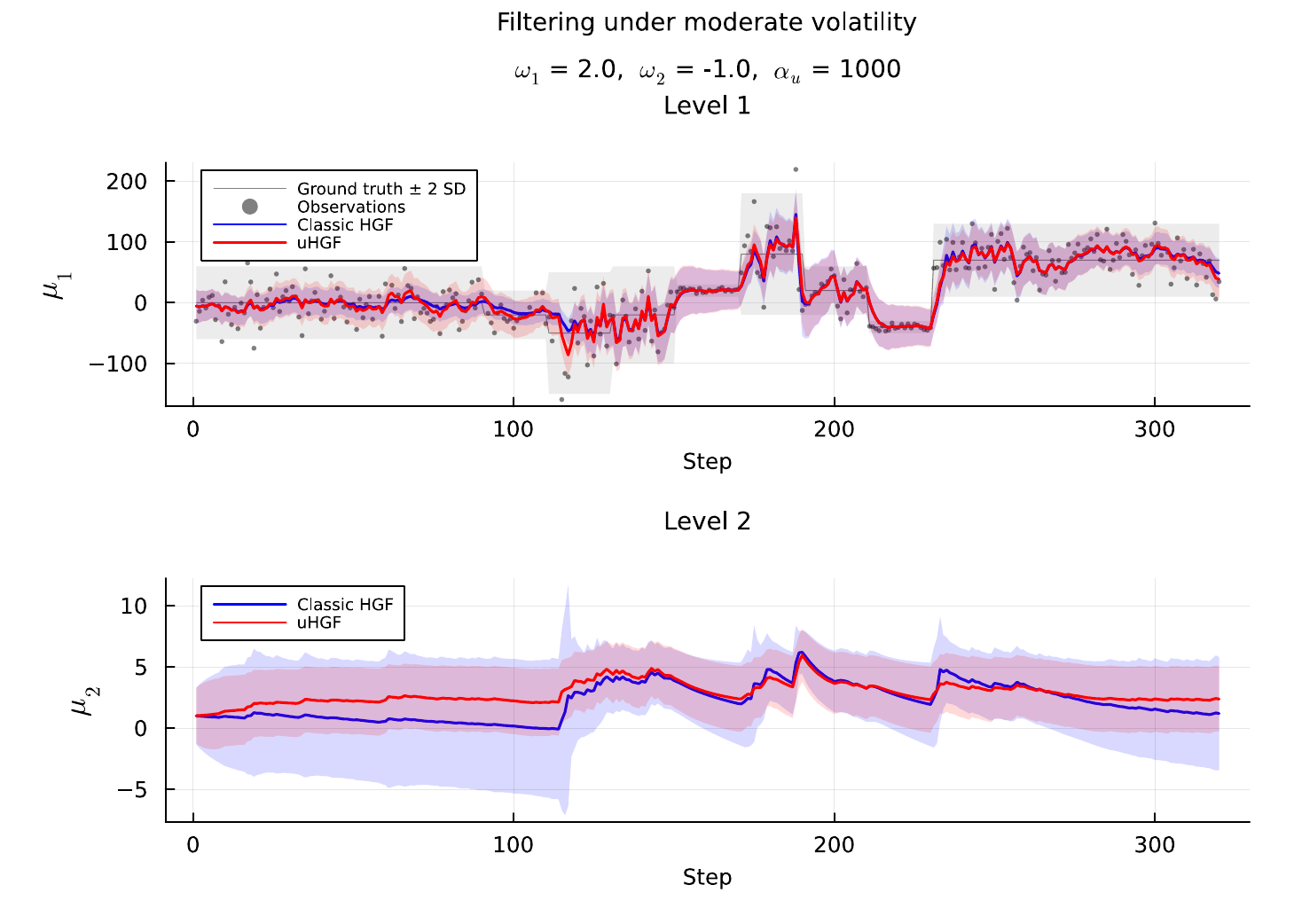}
    \caption{Simulation~2: Under standard conditions ($\omega_1 = 2$, $\omega_2 = -1$), both the classic HGF and the uHGF produce valid posterior trajectories at both levels. Grey shading shows the ground-truth mean $\pm 2$ standard deviations.}
    \label{fig:sim2}
\end{figure}

To confirm that the new equations produce valid filtering results under standard conditions, we filtered a 320-observation time series from \citet{mathys_hierarchical_2020} with parameters $\omega_1 = 2$, $\omega_2 = -1$, and input variance $\alpha_u = 1000$. Both methods completed filtering without incident. Fig.~\ref{fig:sim2} shows the resulting trajectories against the ground-truth mean (grey line) and $\pm 2$ standard deviations (grey shading). At both levels, both methods track the data successfully. The trajectories differ somewhat (level~1: RMSE~$ = 5.1$; level~2: max$\lvert\Delta\mu_2\rvert = 2.3$), reflecting the uHGF's improved quadratic approximation, whose superiority is confirmed by the substantially lower KL divergence in Simulation~1.

\subsubsection{Robustness under extreme prediction errors}

\begin{figure}[ht]
    \centering
    \includegraphics[width=\linewidth]{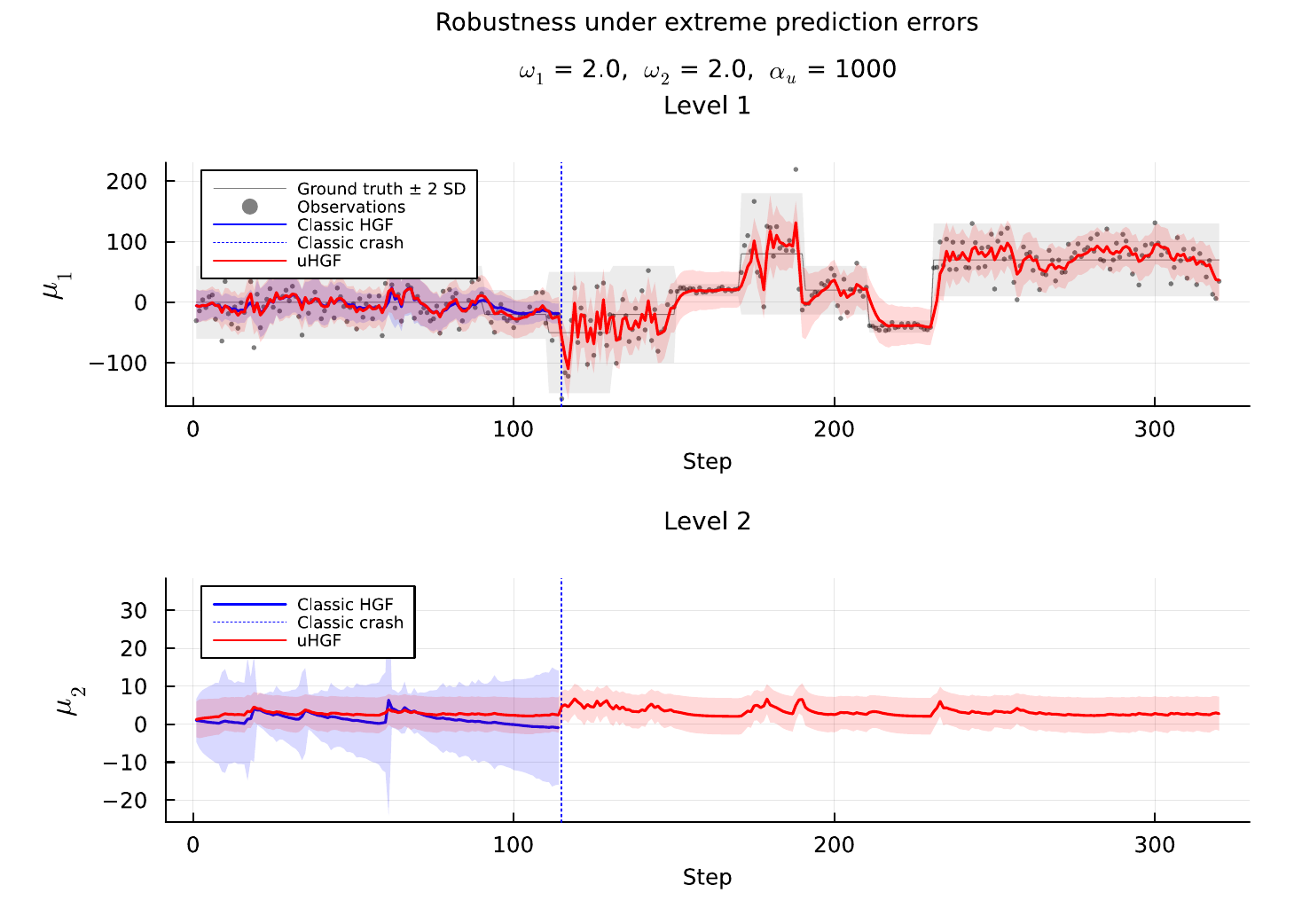}
    \caption{Simulation~3: When confronted with regime changes under high meta-volatility ($\omega_1 = 2$, $\omega_2 = 2$), the classic HGF crashes at step~115 (dotted blue line) due to negative~$\pi_2$. The uHGF completes all 320 steps and adapts to each regime change.}
    \label{fig:sim3}
\end{figure}

In the third simulation, we filtered the same reference time series but now with high meta-volatility ($\omega_2 = 2$), creating conditions where large prediction errors at level~1 propagate forcefully to level~2. As shown in Fig.~\ref{fig:sim3}, the classic HGF crashed at step~115 due to negative posterior precision at level~2. The uHGF, by contrast, completed all 320 steps while maintaining strictly positive precision throughout (min~$\pi_2 = 0.15$). After each regime change in the input, the uHGF adapted its volatility estimate and continued tracking.

\subsubsection{Parameter-space coverage}

\begin{figure}[ht]
    \centering
    \includegraphics[width=0.85\linewidth]{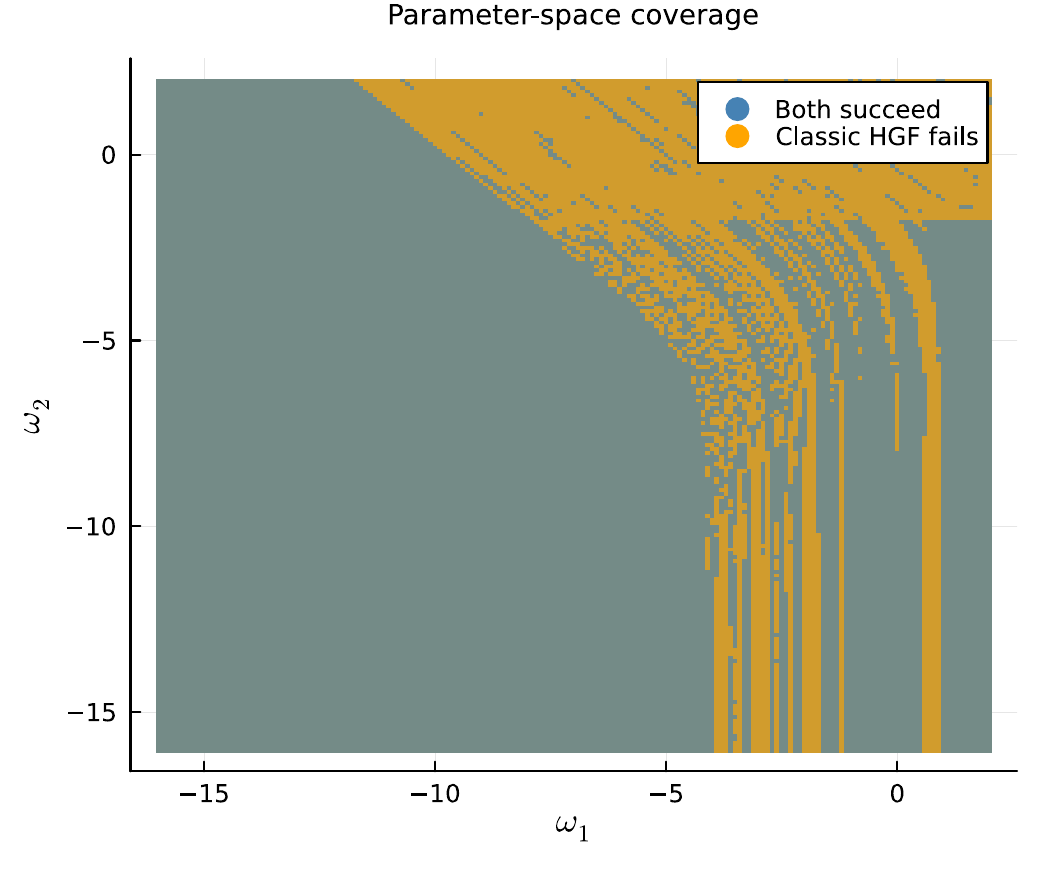}
    \caption{Simulation~4: Parameter-space coverage when filtering the reference time series. Blue indicates parameter sets where both methods succeed; orange indicates where the classic HGF crashes but the uHGF completes. The uHGF extends the usable parameter space to 100\% coverage.}
    \label{fig:sim4}
\end{figure}

Finally, we systematically scanned a fine grid of $181 \times 181 = 32{,}761$ parameter combinations over $\omega_1 \in [-16, 2]$ and $\omega_2 \in [-16, 2]$ (step size 0.1 in both dimensions), with $\kappa_\dagger = 1$ and all other parameters fixed, while filtering the same reference time series. As shown in Fig.~\ref{fig:sim4}, the classic HGF succeeded on 25{,}331 of the 32{,}761 parameter sets (77.3\%), while the uHGF succeeded on all 32{,}761 (100\%). The 7{,}430 classic-only failures (orange) form a large connected region at higher values of $\omega_1$ and $\omega_2$, where the implied step-to-step volatility becomes large enough to trigger extreme prediction errors. The uHGF extends the usable parameter space to full coverage.

\subsection{Extension to the generalized HGF}
\label{sec:ghgf}

The generalized HGF \citep[gHGF;][]{weber_generalized_2025} extends the space of generative models underlying the HGF by introducing \emph{value coupling}, whereby a parent node affects the mean of a child node's predicted value, alongside the existing \emph{volatility coupling}, whereby a parent node affects the child's evolution rate. In the gHGF, a state $x_a$ can simultaneously have value parents $x_{b_i}$ and volatility parents $x_{\check{a}_j}$. Its predicted mean and precision at time step $k$ are
\begin{align}
    \hat{\mu}_a^{(k)} &= \lambda_a \mu_a^{(k-1)} + P_a^{(k)}, \\[0.8em]
    \hat{\pi}_a^{(k)} &= \frac{1}{\dfrac{1}{\pi_a^{(k-1)}} + \Omega_a^{(k)}},
\end{align}
with total predicted drift
\begin{equation}
    P_a^{(k)} = t^{(k)} \!\left( \rho_a + \sum_i \alpha_{b_i,a} \, g_{b_i,a}\!\left(\mu_{b_i}^{(k-1)}\right) \right)
\end{equation}
and total predicted volatility
\begin{equation}
    \Omega_a^{(k)} = t^{(k)} \exp\!\left( \omega_a + \sum_j \kappa_{\check{a}_j,a} \, \mu_{\check{a}_j}^{(k-1)} \right),
\end{equation}
where $\lambda_a$ is the autoconnection parameter, $\rho_a$ the tonic drift, $\alpha_{b_i,a}$ the value coupling strength with coupling function $g_{b_i,a}$, and $\kappa_{\check{a}_j,a}$ the volatility coupling strength \citep[Eqs~34--37]{weber_generalized_2025}.

The precision update equations for value parents are always well-behaved in the linear case: $\pi_b^{(k)} = \hat{\pi}_b^{(k)} + \alpha_{b,a}^2 \hat{\pi}_a^{(k)} > 0$ because both summands are strictly positive \citep[Eqs~38--39]{weber_generalized_2025}. This also applies to nonlinear couplings where the second derivative of the coupling function vanishes almost everywhere (e.g., ReLUs---cf. \citet{weber_generalized_2025}, Eq. 50). The update equations for volatility parents, however, share exactly the functional form given in Section~\ref{sec:theory}: writing the variational energy of volatility parent $x_{\check{a}}$ as a function of its value, the terms $-\frac{1}{2}\log(\sigma_a^0 + \Omega_a^{(k)})$ and $-\frac{1}{2}\beta_a/(\sigma_a^0 + \Omega_a^{(k)})$ depend on $x_{\check{a}}$ through $\Omega_a^{(k)} = t\exp(\omega_a + \kappa_{\check{a},a} x_{\check{a}} + \ldots)$, which has the same exponential structure as before. The prior term $-\frac{1}{2}\hat{\pi}_{\check{a}}(x_{\check{a}} - \hat{\mu}_{\check{a}})^2$ also has the same form, even though $\hat{\mu}_{\check{a}}$ may now include drift contributions from the volatility parent's own value parents.

\paragraph{Single volatility child.}
When the volatility parent $x_{\check{a}}$ has a single volatility child $x_a$, the uHGF approach applies without modification: the variational energy has the same functional form as in Section~\ref{sec:theory}, and the full equations of Section~3.4 carry over directly. The gHGF's richer prediction step (incorporating drift from value parents and contributions from multiple volatility parents) feeds into the predicted quantities $\hat{\mu}$ and $\hat{\pi}$, but once these are computed, the volatility update proceeds exactly as described in Section~3.4.

\paragraph{Multiple volatility children.}
If $x_{\check{a}}$ is volatility parent of several children $x_{a_1}, \ldots, x_{a_n}$, we write $\kappa_i := \kappa_{\check{a},a_i}$ for the coupling strength from $x_{\check{a}}$ to child $x_{a_i}$, and $\tilde{\omega}_{a_i} := \omega_{a_i} + \sum_{j:\,\check{a}_j \neq \check{a}} \kappa_{\check{a}_j,a_i}\,\mu_{\check{a}_j}^{(k-1)}$ for the effective tonic offset of child $x_{a_i}$, absorbing contributions of all its other volatility parents (held fixed under the mean-field approximation). Each child then contributes its own pair of terms to the variational energy:
\begin{equation}
    I(x) = \sum_{i=1}^{n} \left[J_{1,i}(x) + J_{2,i}(x)\right] + J_3(x),
\end{equation}
where $J_{1,i}$ and $J_{2,i}$ are the log-normalizer and scaled-uncertainty terms from child~$a_i$, and $J_3$ is the quadratic prior. Since each $J_{1,i}$ is concave and $J_3$ is quadratic, the concave variational energy $K = \sum_i J_{1,i} + J_3$ is again concave everywhere, and the first expansion's precision
\begin{equation}
    \pi_{L_1} = \hat{\pi}_{\check{a}} + \sum_{i=1}^{n} \frac{\kappa_i^2}{2} w_i (1 - w_i)
\end{equation}
is strictly positive. The mean of the first expansion generalizes to
\begin{equation}
    \mu_{L_1} = \hat{\mu}_{\check{a}} + \frac{1}{\pi_{L_1}} \sum_{i=1}^{n} \frac{\kappa_i}{2}\, w_i\, \delta_i.
\end{equation}

For the remaining expansions, note that each volatility child~$a_i$ contributes its own $J_{2,i}$ term which undergoes a transition from $-\beta_i/(2\sigma_{a_i}^0)$ to $0$ in a region of~$x$ centred at approximately $x \approx (\log\sigma_{a_i}^0 - \log t - \tilde{\omega}_{a_i})/\kappa_i$, with width $O(1/\kappa_i)$. Different volatility children generally have their transition regions at different locations (because they differ in $\tilde{\omega}_{a_i}$, $\kappa_i$, or $\sigma_{a_i}^0$), and each transition can create a secondary mode. Near volatility child~$a_i$'s mode, the other children's $J_{2,j}$ terms ($j \neq i$) are in this case approximately constant --- either in their left-flat or right-flat regime --- and therefore do not affect the mode location. This means the single-child Lambert~$W$ formula of Section~3.4 can be applied independently for each volatility child to find its approximate mode~$x_i^*$.

If two volatility children happen to have nearly identical transition regions (i.e., similar $\gamma_{c,i} = \log t + \kappa_i \hat{\mu}_{\check{a}} + \tilde{\omega}_{a_i}$), then their $J_2$ terms interact and the per-child Lambert~$W$ estimates will be somewhat off. However, this does not threaten the validity of the approximation: the first expansion $L_1$ still has guaranteed positive precision (it does not depend on mode locations at all), the expansions at the Lambert~$W$ points use the full variational energy $I$ including all children's contributions for computing $\pi_{L_{2,i}}$ and $\mu_{L_{2,i}}$, and the softmax-weighted moment matching produces positive precision by construction. The only consequence of overlapping transition regions is a potential decrease in approximation quality.

Specifically, for each volatility child~$a_i$ we compute the Lambert~$W$ mode exactly as in Section~3.4, using that child's parameters $(\kappa_i, \tilde{\omega}_{a_i}, \sigma_{a_i}^0, \beta_i)$ and the parent's prediction $(\hat{\mu}_{\check{a}}, \hat{\pi}_{\check{a}})$:
\begin{equation}
    y_i^* = \gamma_{c,i} - \frac{1}{2\hat{\pi}_{y,i}} + W_0\!\left(\frac{\beta_i}{2\hat{\pi}_{y,i}} \exp\!\left(\frac{1}{2\hat{\pi}_{y,i}} - \gamma_{c,i}\right)\right),
\end{equation}
with $\gamma_{c,i} = \log t + \kappa_i \hat{\mu}_{\check{a}} + \tilde{\omega}_{a_i}$ and $\hat{\pi}_{y,i} = \hat{\pi}_{\check{a}}/\kappa_i^2$. Converting back to the parent's native variable gives $x_i^* = (y_i^* - \log t - \tilde{\omega}_{a_i})/\kappa_i$. At each~$x_i^*$, we construct a quadratic expansion $L_{2,i}$ using the full variational energy~$I$ (including all volatility children's contributions):
\begin{align}
    \pi_{L_{2,i}} &= \hat{\pi}_{\check{a}} + \sum_{j=1}^{n} \frac{\kappa_j^2}{2}\, w_j(x_i^*)\!\left(w_j(x_i^*) + \bigl(2w_j(x_i^*) - 1\bigr)\,\delta_j(x_i^*)\right), \\[0.8em]
    \mu_{L_{2,i}} &= x_i^* + \frac{1}{\pi_{L_{2,i}}} \left(\sum_{j=1}^{n} \frac{\kappa_j}{2}\, w_j(x_i^*)\, \delta_j(x_i^*) - \hat{\pi}_{\check{a}}\,(x_i^* - \hat{\mu}_{\check{a}})\right).
\end{align}
As in Section~3.2, if $\pi_{L_{2,i}} \leq 0$ for some volatility child (because $x_i^*$ falls far from the actual mode), we fall back to the concave curvature $-K''(x_i^*)$, which is always positive.

\paragraph{Softmax blending and moment matching.}

With $n+1$ quadratic expansions ($L_1$ and $L_{2,1}, \ldots, L_{2,n}$), we assign softmax weights based on the variational energy at each expansion's mean:
\begin{equation}
    b_i = \frac{\exp\!\bigl(I(\mu_{L_{2,i}})\bigr)}{\exp\!\bigl(I(\mu_{L_1})\bigr) + \sum_{j=1}^{n} \exp\!\bigl(I(\mu_{L_{2,j}})\bigr)}, \quad b_0 = 1 - \sum_{i=1}^{n} b_i.
\end{equation}
The approximate posterior is then the moment-matched Gaussian of the $(n+1)$-component mixture:
\begin{align}
    \mu_L &= b_0\, \mu_{L_1} + \sum_{i=1}^{n} b_i\, \mu_{L_{2,i}}, \\[0.8em]
    \sigma_L^2 &= b_0 \!\left(\frac{1}{\pi_{L_1}} + (\mu_{L_1} - \mu_L)^2\right) + \sum_{i=1}^{n} b_i \!\left(\frac{1}{\pi_{L_{2,i}}} + (\mu_{L_{2,i}} - \mu_L)^2\right), \\[0.8em]
    \pi_L &= 1/\sigma_L^2.
\end{align}
When the variational posterior is unimodal near~$\hat{\mu}_{\check{a}}$, we have $b_0 \approx 1$ and the result reduces to $L_1$. When a single volatility child~$a_i$ dominates (e.g., because its $\beta_i/\sigma_{a_i}^0$ ratio is much larger), $b_i \approx 1$ and we recover $L_{2,i}$. For $n = 1$, this reduces to the two-component scheme of Sections~3.2 and~3.4.

\paragraph{Simultaneous value and volatility children.}
In the gHGF, a node can be value parent of some children and volatility parent of others at the same time. For linear value coupling, each value child~$b_j$ contributes a quadratic term $-\frac{1}{2}\hat{\pi}_{b_j}\alpha_j^2 t^2(x - c_j)^2$ to $I$, where $c_j$ depends on the child's posterior and prediction. These terms are concave and can be absorbed into an effective prior with increased precision $\tilde{\pi} = \hat{\pi}_{\check{a}} + \sum_j \hat{\pi}_{b_j}\alpha_j^2 t^2$ and appropriately shifted mean $\tilde{\mu}$. The per-child Lambert~$W$ modes, softmax weights, and moment matching then proceed as above with $\tilde{\pi}$ and $\tilde{\mu}$ replacing $\hat{\pi}_{\check{a}}$ and $\hat{\mu}_{\check{a}}$, while the value-child contributions to $\pi_{L_1}$ further increase the guaranteed-positive precision of the first expansion.

In summary, the uHGF makes gHGF networks robust across their entire parameter space: value coupling updates (which were already safe in the linear case), volatility coupling updates (which are now stabilized), and mixed configurations all produce valid, positive posterior precisions.

\section{Discussion}

The update equations introduced in this report solve the problem of negative posterior precision in HGF volatility updates. The solution rests on three ideas. First, when computing the precision of the first quadratic expansion, we use only the concave part of the variational energy, excluding the term $J_2$ which is responsible for the convex region. This guarantees positive precision everywhere. Second, we locate the second mode of the variational energy---when it exists---via a closed-form approximation based on the Lambert~$W$ function \citep{corless_lambert_1996}, avoiding iterative root-finding. Third, we combine the two quadratic expansions into a single Gaussian posterior through softmax-weighted mixture moment matching, which respects the relative heights of the two modes and produces positive precision by construction.

\subsection{Relationship to the original approximation}

The modified update equations reduce to the original ones whenever the variational energy is approximately quadratic, which is the case when the previous posterior uncertainty about the child node ($\sigma_\dagger^0$, or $\alpha$ in canonical form) is on a similar scale to the total posterior uncertainty ($\beta$). This is the regime that predominates under normal filtering conditions with moderate prediction errors, as illustrated by Simulation~2. The difference becomes relevant when a large prediction error pushes $\beta/\alpha$ to high values, creating a convex region or a second mode in the variational energy. In these cases, the original equations either crash (negative precision) or produce a poor Gaussian approximation, while the modified equations continue to track the variational posterior faithfully (Simulations~1, 3, and~4).

\subsection{Implications for parameter estimation}

The robustness of the new equations has direct practical consequences for parameter estimation. When fitting HGF models to behavioural data via maximum-a-posteriori or sampling-based methods, the optimizer explores parameter space, evaluating the likelihood at each candidate parameter set by filtering the entire input sequence. With the original equations, parameter sets that produce even a single instance of negative precision during filtering are invalid, leading to undefined likelihoods. This creates hard boundaries in the objective function that gradient-based optimizers struggle with and that sampling methods must explicitly avoid. The modified equations eliminate these boundaries entirely: every parameter set yields a valid filtering trajectory and therefore a well-defined likelihood. This should make optimization both faster (no wasted evaluations at crashed parameter sets) and more reliable (no risk of optimizers converging to boundary artefacts), and it makes the full parameter space accessible to gradient-based methods.

\subsection{The Lambert $W$ closed form}

The use of the Lambert~$W_0$ function to locate the second mode of the variational energy provides a closed-form alternative to iterative root-finding. This is advantageous for two reasons. First, it is computationally efficient: evaluating $W_0$ by Halley's method converges in very few iterations. Second, it is analytically transparent: the formula (Eq.~\ref{eq:lambertw_mode}) makes explicit how the mode location depends on the prediction error ($\beta$) and the predicted mean ($\gamma$). The approximation is exact in the limit $\alpha \to 0$, which is precisely the regime where the second mode appears. For finite $\alpha$, the residual error is corrected by the Newton step in the second expansion's mean $\mu_{L_2}$.

\subsection{Extension to the generalized HGF}

The extension to the gHGF \citep{weber_generalized_2025} described in Section~\ref{sec:ghgf} shows that the modified equations generalize naturally to networks with multiple children per parent and with mixed value and volatility coupling. The key insight is that the concave structure of $J_1$ and $J_3$ is preserved regardless of the number of children, and that each volatility child's $J_2$ term creates its own transition region in a predictable location. This allows the per-child Lambert~$W$ strategy to find each secondary mode independently. The $(n+1)$-component softmax blending then combines all expansions into a single Gaussian posterior in a principled way.

\subsection{Limitations}

The approach presented here inherits the main limitation of the original HGF: the posterior at each node is approximated by a single Gaussian. When the variational posterior is strongly bimodal, the moment-matched Gaussian may be wider than either mode and centred between them, which is a less informative summary than the full bimodal distribution would be. In practice, however, strong bimodality typically arises only transiently (e.g., immediately following a regime change) and resolves within a few time steps as the filter adapts. A second limitation is that the Lambert~$W$ approximation to the second mode is derived in the limit $\sigma_\dagger^0 \to 0$. For moderate $\sigma_\dagger^0$, the mode location may be slightly off, but this is corrected by the Newton step and has negligible impact on the final moment-matched posterior.

\subsection{Software availability}

A lean Julia reference implementation of the modified update equations is available at \url{https://github.com/ilabcode/UnboundedHGF.jl}. The modifications have also been incorporated into the Julia package \texttt{HierarchicalGaussianFiltering.jl}\footnote{\url{https://github.com/ComputationalPsychiatry/HierarchicalGaussianFiltering.jl}}, the Python package \texttt{pyhgf}\footnote{\url{https://github.com/ComputationalPsychiatry/pyhgf}}, and the MATLAB HGF Toolbox.\footnote{\url{https://github.com/ComputationalPsychiatry/hgf-toolbox}}

\bibliographystyle{apalike}
\bibliography{uhgf}

\end{document}